\documentclass{article}

\usepackage{algorithm}
\usepackage{algorithmic}
\usepackage{epsfig}
\usepackage{subcaption}
\usepackage{calc}
\usepackage{amsfonts}
\usepackage{amssymb}
\usepackage{amstext}
\usepackage{amsmath}
\usepackage{amsthm}
\usepackage{multicol}
\usepackage{pslatex}

\usepackage{graphicx}
\usepackage{tikz}
\usetikzlibrary{positioning}
\usetikzlibrary{arrows}

\usepackage{xcolor}
\usepackage{url}

\newtheorem{definition}{Definition}[section]
\newtheorem{proposition}{Proposition}[section]
\newtheorem{theorem}{Theorem}[section]
\newtheorem{lemma}{Lemma}[section]
\newtheorem{corollary}{Corollary}[section]

\title{Online Learning of Non-Markovian\\ Reward Models}

\author{
Gavin Rens\\{\small KU Leuven, Belgium, email: gavin.rens@kuleuven.be}\\[3mm]
Jean-Fran\c{c}ois Raskin\\ {\small Universit\'e Libre de Bruxelles, Belgium, email: jraskin@ulb.ac.be}\\[3mm]
Rapha\"el Reynouard\\ {\small Universit\'e Libre de Bruxelles, Belgium, email: raphael.reynouard@ulb.ac.be}\\[3mm]
Giuseppe Marra\\ {\small KU Leuven, Belgium, email: giuseppe.marra@kuleuven.be}
}

\begin{document}

\maketitle


\begin{abstract}
\noindent There are situations in which an agent should receive rewards only after having accomplished a series of previous tasks, that is, rewards are {\em non-Markovian}. One natural and quite general way to represent history-dependent rewards is via a {\em Mealy machine}, a finite state automaton that produces output sequences from input sequences. In our formal setting, we consider a Markov decision process (MDP) that models the dynamics of the environment in which the agent evolves and a Mealy machine synchronized with this MDP to formalize the non-Markovian reward function. While the MDP is known by the agent, the reward function is unknown to the agent and must be learned.

Our approach to overcome this challenge is to use Angluin's $L^*$ active learning algorithm to learn a Mealy machine representing the underlying non-Markovian reward machine (MRM). Formal methods are used to determine the optimal strategy for answering so-called membership queries posed by $L^*$.

Moreover, we prove that the expected reward achieved will eventually be at least as much as a given, reasonable value provided by a domain expert. We evaluate our framework on three problems. The results show that using $L^*$ to learn an MRM in a non-Markovian reward decision process is effective.
\end{abstract}

\section{\uppercase{Introduction}}
\label{sec:introduction}

\noindent Traditionally, a Markov Decision Process (MDP) models the probability of going to a state $s'$ from the current state $s$ while taking a given action $a$ together with an {\em immediate reward} that is received while performing $a$ from $s$.
This immediate reward is defined regardless of the history of states traversed in the past. Such immediate rewards thus have the {\em Markovian property}. But many situations  require the reward to depend on the history of states visited so far. A reward may depend on the particular sequence of (sub)tasks that has been completed. For instance, when a nuclear power plant is shut down in an emergency, there is a specific sequence of operations to follow to avoid a disaster; or in legal matters, there are procedures to follow which require documents to be submitted in a particular order.
So, many tasks agents could face need us to reason about rewards that depend on some history, not only the immediate situation (non-Markovian).

Learning and maintaining non-Markovian reward functions is useful for several reasons:
$(i)$ Many tasks are described intuitively as a sequence of sub-tasks or mile-stones, each with their own reward (cf.\ the related work below)
$(ii)$ Possibly, not all relevant features are available in state descriptions, or states are partially observable, making it necessary to visit several states before (more) accurate rewards can be dispensed~\cite{abz10,twkvcm19b}.
$(iii)$ Automata (reward machines) are useful for modeling  desirable and undesirable situations facilitating tracking and predicting beneficial or detrimental situations~\cite{abeknt18,kpr18,dfip19}.
Actually, in practice, it can be argued that non-Markovian tasks are more the norm than Markovian ones.
\begin{figure*}
\centering
\includegraphics[scale=0.4]{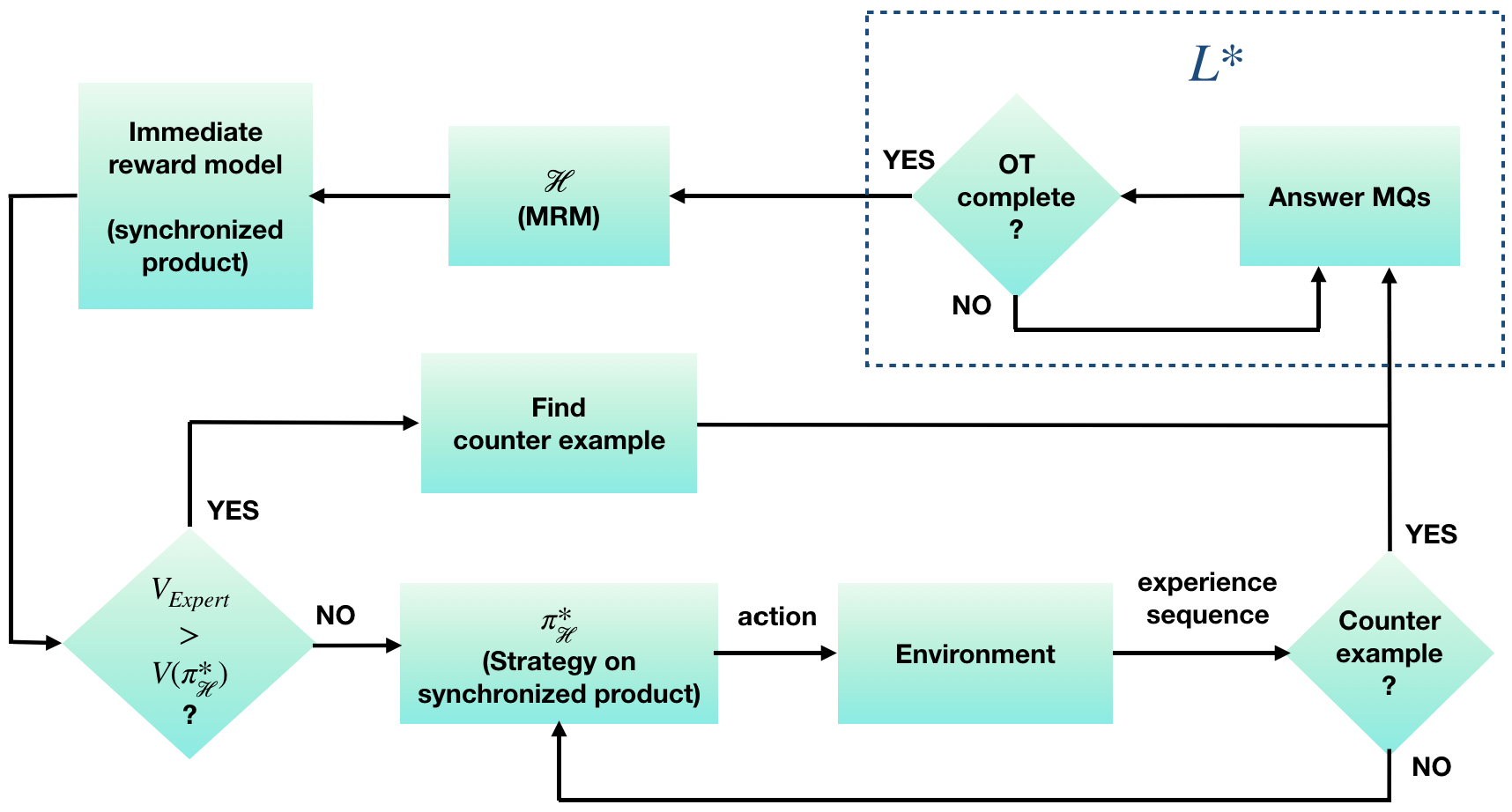}
\caption{
\label{fig:flow-diagram}
Flow diagram of the proposed framework for learning and exploiting Mealy reward machines (MRMs).
}
\end{figure*}

In this work, we assume that the states and the transition function are known, but the reward function is unknown. It is also assumed that the reward behavior of the environment is possibly non-Markovian. The aim is to learn a model of the non-Markovian reward function and use it whenever possible, even if it is not yet known to be a complete model.
We describe an {\em active} learning algorithm to automatically learn non-Markovian reward models by executing `experiments' on the system. In our case, a non-Markovian reward model is represented as a Mealy machine; a deterministic finite automaton (DFA) that produces output sequences that are rewards, from input sequences that are $\langle$state, action$\rangle$ observations. We refer to such finite state reward models as  \textit{Mealy Reward Machines} (MRM).
Our framework (Fig.~\ref{fig:flow-diagram}) is considered to be for online systems, alternating between a learning phase and an exploitation phase. The learning phase is based on Angluin's $L^*$ active learning algorithm \cite{a87} to learn finite automata. It is an \textit{active} learning algorithm because it poses a series of queries to a teacher who must provide answers to an agent performing experiments on/in the system being learnt. 
In our case, the learner is an agent in an MDP and the teacher is the environment. The queries consist of sequences of observations and the answers are the corresponding sequences of rewards that the agent experiences. In this setting, answering membership queries is already a challenge.
An observations is a function of an action and the successor state.
Encountering a particular observation sequence when actions are stochastic and observations have the functional dependencies just mentioned is non-trivial.

When $L^*$ has enough input/output data, it can build a corresponding MRM as an hypothesis of the true underlying MRM. In our framework, whenever an hypothesis is inferred, the agent enters the exploitation phase. The agent then acts to maximize rewards until it encounters a sequence of observations and rewards which contradicts the current hypothesis. The agent then goes back to learning with $L^*$.

Due to the stochasticity of actions in MDPs, a challenge is for the agent to experience exactly the observation sequence posed by $L^*$ as a query, and to do so efficiently. We rely on known formal methods to compute the sequence of actions the agent should execute to encounter the observation sequence with highest probability or least expected number of actions.
Another important aspect of the framework is that, given an (hypothesized) MRM, a synchronized product of the MRM and the MDP is induced so that an \textit{immediate} reward function is available from the induced product MDP. This means that existing (Markovian) MDP planning or reinforcement learning techniques can be employed in the exploitation phase. We use model-based optimization methods as developed by \cite{bk08b}.

Furthermore, in our framework, a special \textit{reset} action is introduced at the time the synchronized product is formed. This action is available at every state of the product MDP such that it takes the system to its initial state. In this way, the product MDP is strongly connected and we can then prove some useful properties of our framework. The engineer needs not be concerned with the reset action when modeling the system, that is, the MDP of the system does not have to mention this action. However, the engineer should keep in mind that the framework is best used for systems which can actually be reset at any time.
Another useful feature of our framework is that it allows a domain expert to provide a value that the system should be able to achieve in the long-run if it follows the optimal strategy. This `expert value' needs not be the optimal expected value, but some smaller value that the system owner wants a guarantee for.

Our contribution is to show how a non-Markovian reward function for a MDP can be actively learnt and exploited to play optimally according to this non-Markovian reward function in the MDP. We provide a framework for completely and correctly learning the underlying reward function with guarantees under mild assumptions. To the best of our knowledge, this is the first work which employs traditional automata inference for learning a non-Markovian reward model in an MDP setting.


\paragraph{Related Work} There has been a surge of interest in non-Markovian reward functions recently, with most papers on the topic being publications since 2017. But unlike our paper, most of those papers are not concerned with {\em learning} non-Markovian reward functions.

An early publication that deserves to be mentioned is~\cite{bbg96}. 
In this paper, the authors propose to encode non-Markovian rewards by assigning values using temporal logic formulae. The reward for being in a state is then the value associated with a formula that is true in that state. 
They were the first to abbreviate the class of MDP with non-Markovian reward as  \textit{decision processes with non-Markovian reward} or NMRDP for short.
A decade later, \cite{tgspk06}
presented the Non-Markovian Reward Decision Process Planner: a software platform for the development and experimentation of methods for decision-theoretic planning with non-Markovian rewards.
In both the cases above, the non-Markovian reward functions is \textit{given} and does not need to be learned.

A closely related and possibly overlapping research area is the use of temporal logic (especially linear temporal logic (LTL)), for specifying tasks in Reinforcement Learning (RL) \cite{abeknt18,tkvm18a,tkvm18b,ctkvm19,dfip19,hak19}.

Building on recent progress in temporal logics over finite traces (LTL$_f$), \cite{bdp18} adopt linear dynamic logic on finite traces (LDL$_f$; an extension of LTL$_f$) for specifying non-Markovian rewards, and provide an automaton construction algorithm for building a Markovian model. The approach is claimed to offer strong minimality and compositionality guarantees.
In those works, the non-Markovian reward function is given and does not have to be learned as in our setting.

In another paper \cite{ccsm18}, the authors are concerned with both the specification and effective exploitation of non-Markovian rewards in the context of MDPs. They specify non-Markovian reward-worthy behavior in LTL. These behaviors are then translated to corresponding deterministic finite state automata whose accepting conditions signify satisfaction of the reward-worthy behavior. These automata accepting conditions form the basis of Markovian rewards by taking the product of the MDP and the automaton (as we do).

None of the research mentioned above is concerned with \textit{learning} non-Markovian reward functions.
However, in the work of \cite{twkvcm19b}, an agent incrementally learns a reward machine (RM) in a partially observable MDP (POMDP). 
They use a set of traces as data to solve an optimization problem. ``If at some point [the RM] is found to make incorrect predictions, additional traces are added to the training set and a new RM is learned.''
%
Their approach is also active learning: If on any step, there is evidence that the current RM might not be the best one, their approach will attempt to find a new one. One strength of their method is that the the reward machine is constructed over a set of propositions, where propositions can be combined to represent transition/rewarding conditions in the machine. Currently, our approach can take only single observations as transition/rewarding conditions.
However, they do not consider the possibility to compute optimal strategies using \textit{model-checking techniques}. 

Moreover, our approach is different to that of \cite{twkvcm19b} in that ours agents are guided by the $L^*$ algorithm to answer exactly the queries required to find the underlying reward machine. The approach of \cite{twkvcm19b} does not have this guidance and interaction with the learning algorithm; traces for learning are collected by random exploration in their approach.

%

Next, we cover the necessary formal concepts and notations.
In Section~\ref{sec:Modeling-Non-Markovian-Rewards}, we define our Mealy Reward Machine (MRM).
Section~\ref{sec:The-Framework} explains how an agent can infer/learn an underlying MRM and present one method for exploiting a learnt MRM.
We discuss the guarantees offered by the framework in Section~\ref{sec:Guarantees}.
Section~\ref{sec:Experimental-Evaluation} reports on experiments involving learning and exploiting MRMs; we consider three scenarios.
The last section concludes this paper and points to future research directions.

\section{\uppercase{Formal Preliminaries}}
\label{sec:Formal-Preliminaries}

\noindent We review Markov Decision Processes (MDPs) and Angluin-style learning of Mealy machines.

An (immediate-reward) MDP is a tuple $\langle S, A, T, R, s_0 \rangle$, where
 \begin{itemize}
 \item  $S$ is a finite set of states,
 \item $A$ is a finite set of actions,
 \item $T:S\times A\times S\mapsto [0,1]$ is the state transition function such that $T(s,a,s')$ is the probability that action $a$ causes a system transition from state $s$ to $s'$,
 \item $R:A\times S\mapsto \mathbb{R}$ is the reward function such that $R(a,s)$ is the immediate rewards for performing action $a$ in state $s$, and
 \item $s_0$ the initial state the system is in.
 \end{itemize}
A non-rewarding MDP (nrMDP) is a tuple $\langle S, A, T, s_0 \rangle$ without a reward function.

The following description is from \cite{v17}.
\cite{a87} showed that finite automata can be learned using the so-called membership and equivalence queries.
Her approach views the learning process as a game between a \textit{minimally adequate teacher} (MAT) and a learner who wants to discover the automaton. In this framework, the learner has to infer the input/output behavior of an unknown Mealy machine by asking queries to a teacher. The MAT knows the Mealy machine $\mathcal{M}$.
Initially, the learner only knows the inputs $\mathcal{I}$ and outputs $\mathcal{O}$ of $\mathcal{M}$. The task of the learner is to learn $\mathcal{M}$ through two types of queries:
(1)
With a \textit{membership query} (MQ), the learner asks what the output is in response to an input sequence
$\sigma \in \mathcal{I}^*$. The teacher answers with output sequence $\mathcal{M}(\sigma)$.
(2)
With an \textit{equivalence query} (EQ), the learner asks whether a hypothesized Mealy machine $\mathcal{H}$ with inputs $\mathcal{I}$ and outputs $\mathcal{O}$ is correct, that is, whether $\mathcal{H}$ and $\mathcal{M}$ are equivalent ($\forall \sigma\in \mathcal{I}^*, \mathcal{M}(\sigma) = \mathcal{H}(\sigma)$). The teacher answers \textit{yes} if this is the case. Otherwise she answers \textit{no} and supplies a \textit{counter-example} $\sigma' \in \mathcal{I}^*$ that distinguishes $\mathcal{H}$ and $\mathcal{M}$ (i.e., such that $\mathcal{M}(\sigma') \neq \mathcal{H}(\sigma')$).

The $L^*$ algorithm incrementally constructs an \textit{observation table} with entries being elements from $\mathcal{O}$.
Two crucial properties of the observation table allow for the construction of a Mealy machine \cite{v17}: closedness and consistency.
We call a closed and consistent observation table \textit{complete}.

\cite{a87} proved that her $L^*$ algorithm is able to learn a finite state machine (incl. a Mealy machine) by asking a polynomial number of membership and equivalence queries (polynomial in the size of the corresponding minimal Mealy machine equivalent to $\mathcal{M}$).
Let $|\mathcal{I}|$ be the size of the input alphabet (observations), $n$ be the total number of states in the target Mealy machine, and $m$ be the maximum length of the counter-example provided by the MAT for learning the machine. Then the correct machine can be produced by asking maximum $O(|\mathcal{I}|^2 + |\mathcal{I}|mn^2)$ queries (using, e.g., the ${L_M}^+$ algorithm) \cite{sg09}.

In an ideal (theoretical) setting, the agent (learner) would ask a teacher whether $\mathcal{H}$ is correct (equivalence query), but in practice, this is typically not possible \cite{v17}.
``Equivalence query can be approximated using a conformance testing (CT) tool \cite{ly96} via a finite number of test queries (TQs). A test query asks for the response of the [system under learning] to an input sequence, similar to a membership query. If one of the test queries exhibits a counter-example then the answer to the equivalence query is \textit{no}, otherwise the answer is \textit{yes}'' \cite{v17}. A finite and complete conformance test suite does exist if we assume a bound on the number of states of a Mealy machine \cite{ly96}.

Our present framework, however, does not rely on conformance testing by performing a particular suite of TQs. Rather, if it is found that the current hypothesis would under-perform compared to what a domain expert expects, the framework executes a weaker kind of conformance testing: the agent performs actions uniformly at random until a counter-example is found.

\section{\uppercase{Modeling Non-Markovian Rewards}}
\label{sec:Modeling-Non-Markovian-Rewards}

\begin{figure}
\centering
\begin{subfigure}[b]{.4\textwidth}
\centering
\includegraphics[width=1\linewidth]{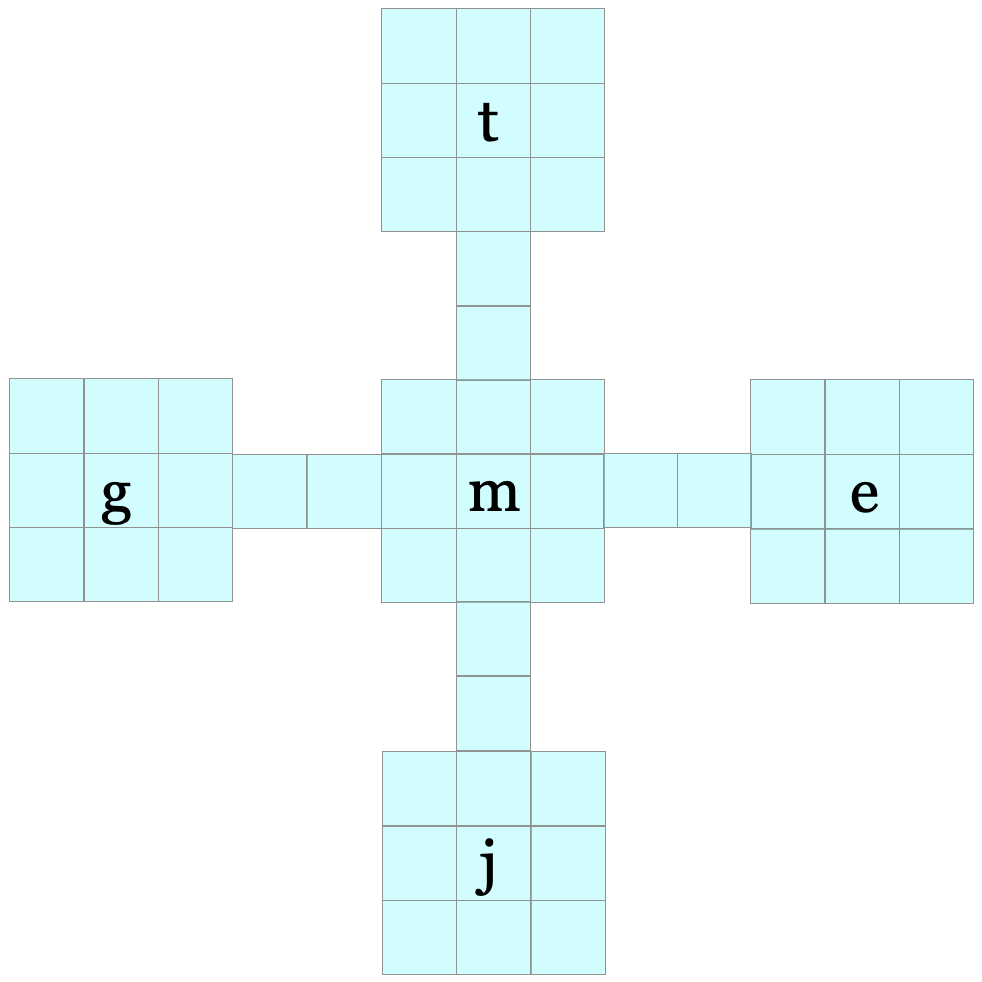}  
  \caption{The treasure-map world. Blank cells contain observation $null$ by default.}
  \label{fig:tmw-scenario}
\end{subfigure}
\hspace{2mm}
\begin{subfigure}[b]{.5\textwidth}
\centering
\begin{tikzpicture}[->,>=stealth',shorten >=1pt,auto,node distance=3cm,
  thick,main node/.style={circle,fill=blue!10,draw,
  font=\sffamily\normalsize\bfseries,minimum size=8mm, scale=0.6, transform shape}]

  \node[main node] (0) {0};
  \node[main node] (1) [right of=0] {1};
  \node[main node] (2) [above right of=1] {2};
  \node[main node] (3) [below right of=1] {3};
  \node[main node] (4) [below right of=2] {4};
  
  \path[every node/.style={near start, font=\sffamily\tiny,fill=none, inner sep=1pt}]
    (0) edge [] node[above=2pt, midway] {$m\mid10$} (1)
    (1) edge [] node[midway] {$e\mid80$} (2)
    edge [] node[left=3pt, midway] {$g\mid70$} (3)
    (2) edge [] node[] {$t\mid80$} (4)
    (3) edge [] node[right=2pt, near start] {$t\mid95$} (4)
    (4) edge [] node[above=2pt, midway] {$j\mid180$} (1);
  \end{tikzpicture}
\vspace{4mm}
\caption{A Mealy reward machine. Self-loops are not shown.}
  \label{fig:tmw-MRM}
\end{subfigure}
\vspace{2mm}
\caption{The treasure-map scenario.}
\label{fig:treasure-map-world}
\end{figure}
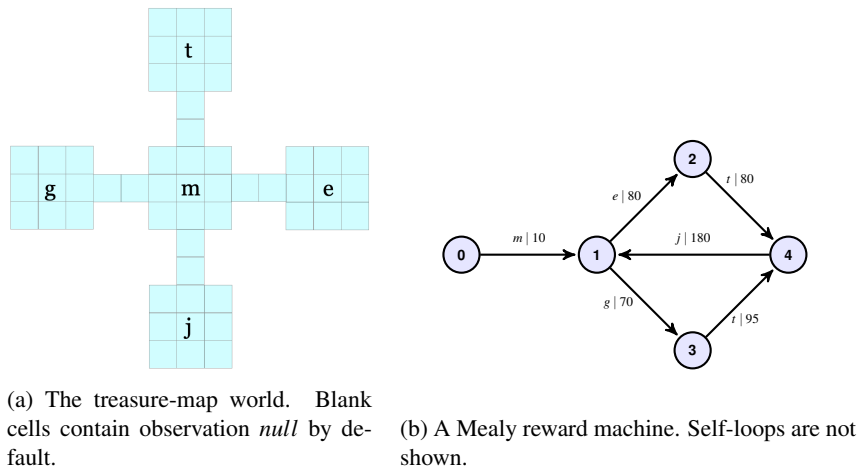

\subsection{Running Example}
\noindent Consider, as a running example (Fig.~\ref{fig:treasure-map-world}), an agent who might stumble upon a person with a map (\textsf{m}) for a hidden treasure (\textsf{t}) and some instruction on how to retrieve the treasure. The instructions imply that the agent purchase some specialized equipment (\textsf{e}) before going to the cave marked on the treasure map. Alternatively, the agent may hire a guide (\textsf{g}) who already has the necessary equipment. If the agent is lucky enough to find some treasure, the agent must sell it at the local jewelry trader (\textsf{j}). The agent can then restock its equipment or re-hire a guide, get some more treasure, sell it and so on. Unfortunately, the instructions are written in a coded language which the agent cannot read. However, the agent recognizes that the map is about a hidden treasure, and thus spurs the agent on to start treasure hunting to experience rewards and learn the missing information.

Besides the four movement actions, the agent can also buy, sell and collect. In the next subsection, we define a labeling function which takes an action and a state, and maps to an observation. For this example, the labeling function has the following effect. If the agent is in the state containing \textsf{m}, \textsf{e} or \textsf{g} and the does the \textit{buy} action, then the agent observes \textsf{m}, \textsf{e} respectively \textsf{g}. If the agent is in the state containing \textsf{t}, and does the \textit{collect} action, then the agent observes \textsf{t}. If the agent is in the state containing \textsf{j}, and does the \textit{sell} action, then the agent observes \textsf{j}. In all other cases, the agent observes $null$.

After receiving the treasure map, the agent might not find a guide or equipment. In general, an agent might not finish a task or finish only one version of a task or subtask. In this world, the agent can reset its situation at any time; it can receive a map again and explore various trajectories in order to learn the entire task with all its variations.

The reward behavior in this scenario is naturally modeled as an automaton where the agent receives a particular reward given a \textit{particular sequence} of observations. There is presumably higher value in purchasing equipment for treasure hunting only \textit{after} coming across a treasure map and thus deciding to look for a treasure. There is more value in being at the treasure with the right equipment than with no equipment, etc.

We shall interpret features of interest as observations: (obtaining) a map, (purchasing) equipment, (being at) a treasure, (being at) a jewelry trader, for example.
Hence, for a particular sequence of input observations, the Mealy machine outputs a corresponding sequence of rewards. If the agent tracks its observation history, and if the agent has a correct automaton model, then it knows which reward to expect (the last symbol in the output) for its observation history as input. 
Figure~\ref{fig:tmw-scenario} depicts the scenario in two dimensions.
The underlying Mealy machine could be represented graphically as in Figure~\ref{fig:tmw-MRM}. It takes observations as inputs and supplies the relevant outputs as rewards.
For instance, if the agent sees $\textsf{m}$, (map), then $\textsf{j}$, then $\textsf{t}$, then the agent will receive rewards 10, then 0 and then 0. And if it sees the sequence $\textsf{m} \cdot\textsf{g} \cdot\textsf{t} \cdot \textsf{j}$, then it will receive the reward sequence $10 \cdot 70\cdot 95\cdot 180$.

We define \textit{intermediate states} as states that do not signify a significant progress towards completion of a task. In Figure~\ref{fig:tmw-scenario}, all blank cells represent intermediate states.
We assume that there is a default reward/cost an agent gets for entering intermediate states. This default reward/cost is fixed and denoted $c$ in general discussions. Similarly, the special null observation ($null$) is observed in intermediate states. An agent might or might not be designed to recognize intermediate states. 
If the agent could not distinguish intermediate states from `significant' states, then $null$ would be treated exactly like all other observations and it would have to learn transitions for $null$ observations (or worse, for all observations associated with intermediate states). If the agent \textit{can} distinguish intermediate states from `significant' states (as assumed in this work), then we can design methods to take advantage of this background knowledge.

\subsection{Mealy Reward Machines}
\noindent We introduce the \textit{Mealy reward machine} to model non-Markovian rewards. These machines take a set $Z$ of observations representing high-level features that the agent can detect (equivalent to the set of input symbols for $L^*$). A labeling function $\lambda:A\times S \mapsto Z\uplus \{null\}$ maps action-state pairs onto observations; $S$ is the set of nrMDP states and $Z$ is a set of observations. The meaning of $\lambda(a,s)= z$ is that $z$ is observed in state $s$ reached via action $a$. The $null$ observation always causes the trivial transition in the machine (i.e. self-loop) and produced a default reward denoted $c$. For Mealy reward machines, the output symbols for $L^*$ are equated with rewards in $\mathbb{R}$.
\begin{definition}[Mealy Reward Machine]
\label{def:RT-Rens}
Given a set of states $S$, a set of actions $A$ and a labeling function $\lambda$, a \emph{Mealy Reward Machine} (MRM) is a tuple $\langle U, u_0, Z, \delta_u, \delta_r, c \rangle$, where
\begin{itemize}
\item $U$ is a finite set of MRM nodes,
\item $u_0 \in U$ is the start node,
\item $Z\uplus \{null\}$ is a set of observations,
\item $\delta_u:U\times Z \mapsto U$ is the transition function, such that $\delta_u(u_i, \lambda(a,s))=u_j$ for $a\in A$ and $s\in S$, specifically, $\delta_u(u_i, null)=u_i$,
\item $\delta_r:U\times Z \mapsto \mathbb{R}$ is the reward-output function, such that $\delta_r(u_i, \lambda(a,s)) = r'$ for $r'\in\mathbb{R}$, $a\in A$ and $s\in S$, specifically, $\delta_r(u_i, null)=c$.
\end{itemize}
We may write $\delta_u^\mathcal{R}$ and $\delta_r^\mathcal{R}$ to emphasize that the functions are associated with MRM $\mathcal{R}$.
\end{definition}
A Markov Decision Process with a Mealy reward machine is thus a non-Markovian reward decision process (NMRDP).
In the diagrams of MRMs, an edge from node $u_i$ to node $u_j$ labeled $ z\mid r $ denotes that $\delta_u(u_i,z)=u_j$ and $\delta_r(u_i,z)=r$. 

In the following definitions, let $s_i\in S$, $a_i\in A$, $z_i\in Z$, and $r_i\in\mathbb{R}$.
\begin{itemize}
\item \emph{interaction trace} of length $k$ in an MDP represents an agent's (recent) interactions with the environment. It has the form $s_0 a_0 r_1 s_1 a_1 r_2 \cdots s_{k-1} a_{k-1} r_{k}$ and is denoted $\sigma_\mathit{inter}$. That is, if an agent performs an action at time $t$ in a state at time $t$, then it reaches the next state at time $t+1$ where/when it receives a reward.
\item An \textit{observation trace} is extracted from an interaction trace (employing a labeling function) and is taken as input to an MRM. It has the form $z_1 z_2 \cdots z_k$ and is denoted $\sigma_z$.
\item A \textit{reward trace} is either extracted from an interaction trace or is given as output from an MRM. It has the form $r_1 r_2 \cdots r_{k}$ and is denoted $\sigma_r$.
\item A \textit{history} has the form $s_0 a_0 s_1 a_1 \cdots s_{k}$ and is denoted $\sigma_h$.
\end{itemize} 

Given a history $\sigma_h$, we extend $\delta_u$ to take histories by defining $\delta^*_{u}(u_i,\sigma_h)$ inductively as 
\[
\delta_u(u_i, \lambda(a_0,s_1))\cdot\delta^*_{u}(\delta_u(u_i, \lambda(a_0,s_1)),a_1 s_2 \cdots s_k).
\]

Finally, we extend $\delta_r$ to take histories by defining $\delta^*_{r}(u_i,\sigma_h)$ inductively as 
\[
\delta_r(u_i, \lambda(a_0,s_1))\cdot\delta^*_{r}(\delta_u(u_i, \lambda(a_0,s_1)),a_1 s_2 \cdots s_k).
\]
$\delta^*_{r}$ explains how an MRM assigns rewards to a history in an MDP.


\subsection{Expected Value of a Strategy under an MRM}
\noindent A (deterministic) strategy to play in a nrMDP $M = \langle S, A, T, s_0 \rangle$ is a function $\pi:S^* \mapsto A$ that associates to all sequences of states $\sigma\in S^*$ of $M$ the action $\pi(\sigma)$ to play.

In this version of the framework, the agent tries to maximize \textit{mean payoff}, defined as
$
\mathcal{MP}(\sigma_r) = \frac{1}{k}\sum_{i=1}^k r_i.
$, where $r_i$ is the reward received at step $i$.
Let $\mathcal{MP}(\mathcal{R})$ be the mean payoff of an infinite reward trace generated by reward model $\mathcal{R}$.
Then the \textit{expected} mean payoff under strategy $\pi$ played in MDP $M$ from state $s$ is denoted as $\mathbb{E}^{M,\pi}_{s}(\mathcal{MP}(\mathcal{R}))$.




Being able to produce a traditional, immediate reward MDP from a non-Markovian rewards decision process is clearly beneficial: One can then apply all the methods developed for MDPs to the underlying NMRDP, whether to find optimal or approximate solutions. The following definition is the standard way to produce an MDP from a non-reward MDP (nrMDP) and a deterministic finite automaton.

\begin{definition}[Synchronized Product]
Given an nrMDP $M = \langle S, A, T, s_0 \rangle$, a labeling function $\lambda:S\mapsto Z\uplus\{null\}$ and an MRM $\mathcal{R}=\langle U, u_0, Z, \delta_u, \delta_r \rangle$, the synchronized product of $M$ and $\mathcal{R}$ under $\lambda$ is defined as an (immediate reward) MDP $P = M \otimes_\lambda \mathcal{R} = \langle S^\otimes, A^\otimes, T^\otimes, R^\otimes, s_0^\otimes \rangle$, where
\begin{itemize}
\item $S^\otimes = S\times U$,
\item $A^\otimes = A$,
\item $T^\otimes((s,u), a, (s',u'))=\left\lbrace
\begin{array}{cl}
T(s, a, s') & \mbox{if } u' = \delta_u(u,\lambda(a,s')) \\
0 & \mbox{otherwise}.
\end{array}
\right.$
\item $R^\otimes(a, (s,u)) = \delta_r(u,\lambda(a,s))$, 
\item $s_0^\otimes = (s_0,u_0)$
\end{itemize}
\end{definition}
Due to MRMs having deterministic transitions ($\delta_u$ and $\delta^*_{u}$ are functions), we have that the histories in $M$ and $P$ are in bijection. This is stated formally in the following proposition. Let $\mathfrak{H}(M)$ and $\mathfrak{H}(P)$ denote all histories that can be generated from $M$, resp.\, $P$.
\begin{proposition}
Let $B:\mathfrak{H}(M)\mapsto\mathfrak{H}(P)$ be defined for $\sigma_h=s_0 a_0 s_1 a_1 \cdots s_{k}$ as
\begin{align*}
B(\sigma_h) = &\quad (s_0, u_0) \\
&\quad a_0 (s_1,\delta_u(u_0,\lambda(a_0,s_1)) \\
&\quad a_1 (s_2, \delta_u(\delta_u(u_0,\lambda(a_0,s_1)),\lambda(a_1,s_2))) \\
&\quad \cdots \\
&\quad a_{k-1} (s_k, last(\delta^*_{u}(u_0,\sigma_h))),
\end{align*}
\end{proposition}
As a consequence, given a strategy $\pi$ for $M$, we can define a strategy $\pi^\otimes$ for $P$ as follows.
For all $\sigma_h\in \mathfrak{H}(M)$, $\pi^\otimes(B(\sigma_h)) = \pi(\sigma_h)$.
\begin{corollary}
The strategies in $M$ and $P$ are in bijection.
\end{corollary}
The following proposition states that the expected value of an nrMDP with an MRM is equal to the expected value of their product under the same strategy.
\begin{proposition}
\label{prp:expected-val-of-product-MDP}
Given an nrMDP $M = \langle S, A, T, s_0 \rangle$, a labeling function $\lambda:S\mapsto Z\uplus\{null\}$, an MRM $\mathcal{R}=\langle U, u_0, Z, \delta_u, \delta_r \rangle$, $P = M \otimes_\lambda \mathcal{R} $, for all strategies $\pi$ for $M$ and $\pi^\otimes$ in bijection, we have that
\[
\mathbb{E}^{M,\pi}_{s_0}(\mathcal{MP}(\mathcal{R})) = \mathbb{E}^{P,\pi^\otimes}_{(s_0,u_0)}(\mathcal{MP}(R^\otimes)).
\]
\end{proposition}

\subsection{Adding a Reset Action}
\noindent We make the important assumption that the environment and the agent's state can be reset. Resetting also sets the underlying and hypothesized MRMs to their start nodes.
Having a reset action available means that the underlying domain need not be strongly connected, that is, there needs not exists a path between all pairs of states.
Having the option to reset at any time allows one to learn in terms of episodes: either an episode ends naturally (when the maximum episode length is reached or when the task is achieved to satisfaction) or the episode is cut short by resetting the system due to lack of progress.
Of course, the agent retains the hypothesized MRM learnt so far. Resetting a system is not always feasible. We are, however, interested in domains where an agent can be placed in a position to continue learning or repeat its task.

The results discussed in the next section rely on the implicit presence of a special reset action $\curvearrowleft$. Instead of modifying the learning procedure, this amounts to adding $\curvearrowleft$ when defining sybchronized products ( $\otimes_\lambda^\curvearrowleft$ ).

Formally, this is done as follows.
Let $s_0$ and $u_0$ be the unique starting state of MDP $M$ and the unique starting node of MRM $\mathcal{R}$, respectively. Then $M \otimes_\lambda^\curvearrowleft \mathcal{R}$ is defined as before, and with $\curvearrowleft$ available at every product state $(s,u)$ such that $T((s,u), \curvearrowleft, (s_0,u_0)) = 1$.
By adding $\curvearrowleft$ in this way, the products become \textit{strongly connected}.

\begin{definition}[Synchronized Product with Homing]
Given an nrMDP $M = \langle S, A, T, s_0 \rangle$, a labeling function $\lambda:S\mapsto Z\uplus\{null\}$, a reset cost $c_\curvearrowleft$ and an MRM $\mathit{\mathcal{R}}=\langle U, u_0, Z, \delta_u, \delta_r \rangle$, we define the homing/`resetable' synchronized product of $M$ and $\mathcal{R}$ under $\lambda$ as an (immediate reward) MDP $P = M \otimes_\lambda^\curvearrowleft \mathcal{R} = \langle S^{\otimes^\curvearrowleft}, A^{\otimes^\curvearrowleft}, T^{\otimes^\curvearrowleft}, R^{\otimes^\curvearrowleft}, s_0^{\otimes^\curvearrowleft} \rangle$, where
\begin{itemize}
\item $S^{\otimes^\curvearrowleft} = S\times U$
\item $A^{\otimes^\curvearrowleft} = A\uplus \{\curvearrowleft\}$
\item $T^{\otimes^\curvearrowleft}((s,u), a, (s',u')) = \left\lbrace
\begin{array}{cl}
T(s, a, s') & \mbox{if } a\neq\curvearrowleft \land u' = \delta_u(u,\lambda(a,s')) \\
0 & \mbox{if } a\neq\curvearrowleft \land u' \neq \delta_u(u,\lambda(a,s'))\\
1 & \mbox{if } a = \curvearrowleft \land (s',u')=(s_0,u_0)
\end{array}
\right.$
\item $R^{\otimes^\curvearrowleft}(a, (s,u)) = \left\lbrace
\begin{array}{cl}
\delta_r(u,\lambda(a,s)) & \mbox{if } a\neq\curvearrowleft\\
c_\curvearrowleft & \mbox{otherwise}
\end{array}
\right.$
\item $s_0^{\otimes^\curvearrowleft} = (s_0,u_0)$
\end{itemize}
\end{definition}
Note that adding the reset action does not change the result of Proposition~\ref{prp:expected-val-of-product-MDP}.

The following proposition states that the expected value of an nrMDP with an MRM is equal to the expected value of their product under the same strategy.
\begin{corollary}
\label{cor:expected-val-of-product-MDP}
Given an nrMDP $M = \langle S, A, T, s_0 \rangle$, a labeling function $\lambda:S\mapsto Z\uplus\{null\}$, an MRM $\mathcal{R}=\langle U, u_0, Z, \delta_u, \delta_r \rangle$, $P = M \otimes_\lambda^\curvearrowleft \mathcal{R} $, for all strategies $\pi$ for $M$ and $\pi^{\otimes^\curvearrowleft}$ in bijection, we have that
\[
\mathbb{E}^{M,\pi}_{s_0}(\mathcal{MP}(\mathcal{R})) = \mathbb{E}^{P,\pi^{\otimes^\curvearrowleft}}_{(s_0,u_0)}(\mathcal{MP}(R^{\otimes^\curvearrowleft})).
\]
\end{corollary}

A strategy $\pi:S^*\mapsto A$ is \textit{memoryless} if $\forall \sigma_1,\sigma_2\in S^*$ such that $last(\sigma_1) = last(\sigma_2)$, then $\pi(\sigma_1) = \pi(\sigma_2)$, that is, if the action played by $\pi$ for sequence $\sigma$ depends only on the last state in $\sigma$.
Because memoryless strategies are sufficient to play optimally in order to maximize $\mathbb{E}^{P,\pi^{\otimes^\curvearrowleft}}_{(s_0,u_0)}(\mathcal{MP}(R^{\otimes^\curvearrowleft}))$ in immediate reward MDPs, and together with Corollary~\ref{cor:expected-val-of-product-MDP}, we conclude that we can play optimally in $M$ under $\mathcal{R}$ in a finite memory strategy $\pi^*$ (finite in the memory required for $\mathcal{R}$; but memoryless if viewed as $P = M \otimes_\lambda^\curvearrowleft \mathcal{R}$). Strategy $\pi^*$ can be computed with, for instance, the STORM probabilistic model checking software package.\footnote{STORM can be found at http://www.stormchecker.org} In this work, we use STORM to compute $\pi^*$ for $M \otimes_\lambda^\curvearrowleft \mathcal{H}$ (product of the nrMDP and the hypothesis MRM).

\section{\uppercase{The Framework}}
\label{sec:The-Framework}

\subsection{Flow of Operation}
\label{sec:Flow-of-Operation}

\noindent A high-level description of the process follows (recall Fig.~\ref{fig:flow-diagram}).
\begin{enumerate}
\item Play a finite number of episodes to answer membership queries by extracting reward traces from the appropriate interaction traces until the observation table (OT) is complete;
\item As soon as OT becomes complete, construct new hypothesized MRM $\mathcal{H}$ from OT and compute the optimal strategy $\pi^*_\mathcal{H}$ for $M \otimes_\lambda^\curvearrowleft \mathcal{H}$ with value $V(\pi^*_\mathcal{H})$ employing STORM; 
\item If $V(\pi^*_\mathcal{H})$ is less than the value provided by a domain expert, then we know that $\mathcal{H}$ must be wrong, in which case, we seek an interaction trace which is a counter example to $\mathcal{H}$ by playing actions uniformly at random (strategy $\pi_u$);
\item Else, if $V(\pi^*_\mathcal{H})$ is greater than or equals the value provided by the domain expert, then repeatedly execute actions using $\pi^*_\mathcal{H}$;
\item Whenever a counter-example to $\mathcal{H}$ is discovered (experienced), stop exploitation and go to step 1 (the learning phase).
\end{enumerate}
\noindent
 Algorithm~\ref{alg:online-active-learning} is the high-level algorithm implementing these steps.

 \begin{algorithm}[h!]
 \begin{normalsize}
 \caption{Approximate Active Learning
 \label{alg:online-active-learning}}
 \begin{algorithmic}
 \STATE Initialize observation table $OT$\;
 \WHILE{alive}
 	\IF{$OT$ \textnormal{is not complete}}
 		\STATE $\sigma_z \gets$ getMQ$(OT)$\;
 		\STATE $\sigma_r \gets$ getExperience$(\sigma_z, s_0)$\;
 		\STATE resolveMQ$(OT, \sigma_z, \sigma_r)$\;
 	\ELSE
 		\STATE $\mathcal{H} \gets$ buildRewardMachine$(OT)$\;
 		\STATE $P = \langle S^\otimes, A^\otimes, T^\otimes, R^\otimes, (s_0,u_0) \rangle \gets M \otimes_\lambda^\curvearrowleft \mathcal{H}$\;
 		\STATE $\pi^*_\mathcal{H}\gets Plan(P)$\;
 		\STATE $s_\mathit{cur} \gets s_0; u_\mathit{cur} \gets u_0$\;
 		\IF{$V(\pi^*_\mathcal{H})<V_\mathit{expert}$}
 		    \REPEAT
 		        \STATE $a \gets random(A^\otimes)$\;
 		        \STATE $(s_\mathit{nxt},u_\mathit{nxt}) \thicksim T((s_\mathit{cur},u_\mathit{cur}), a, \cdot)$\;
     			\STATE $r\gets R^\otimes(a, (s_\mathit{nxt}, u_\mathit{cur}))$\;
     			\STATE Update $\sigma_\mathit{inter}$ with $(a, s_\mathit{nxt}, r)$\;
     			\STATE $s_\mathit{cur} \gets s_\mathit{nxt}; u_\mathit{cur} \gets u_\mathit{nxt}$\;
 		    \UNTIL{$\sigma_\mathit{inter}$ \textnormal{is a counter example to} $\mathcal{H}$}
 		\ELSE
     		\REPEAT
     		    \STATE reset environment and refresh $\sigma_\mathit{inter}$\;
     		    \STATE $s_\mathit{cur} \gets s_0; u_\mathit{cur} \gets u_0$\;
     		    \STATE $steps = 0$\;
     		    \WHILE{$steps < EpisodeLength$}
     		        \STATE increment $steps$ by 1\;
         			\STATE $a \gets \pi^*_\mathcal{H}(s_\mathit{cur},u_\mathit{cur})$
         			\STATE $(s_\mathit{nxt},u_\mathit{nxt}) \thicksim T((s_\mathit{cur},u_\mathit{cur}), a, \cdot)$\;
         			\STATE $r\gets R^\otimes(a, (s_\mathit{nxt}, u_\mathit{cur}))$\;
         			\STATE Update $\sigma_\mathit{inter}$ with $(a, s_\mathit{nxt}, r)$\;
         			\STATE $s_\mathit{cur} \gets s_\mathit{nxt}; u_\mathit{cur} \gets u_\mathit{nxt}$\;
         		\ENDWHILE
     		\UNTIL{$\sigma_\mathit{inter}$ \textnormal{is a counter example to} $\mathcal{H}$}
     	\ENDIF
         \STATE $\sigma_z\gets$ extractObsTrace$(\sigma_\mathit{inter})$\;
         \STATE $\sigma_r\gets$ extractRewTrace$(\sigma_\mathit{inter})$\;
         \STATE addCounterExample$(OT, \sigma_z, \sigma_r)$\;
 	\ENDIF
 \ENDWHILE
 \end{algorithmic}
 \end{normalsize}
 \end{algorithm}

 In the algorithm, ``alive'' implies that there is no specific stopping criterion. In practice, the user of the system/agent will specify when the system/agent should stop collecting rewards.
 Step 1 corresponds to procedures getMQ$(OT)$, getExperience$(\sigma_z, s_0)$ and resolveMQ$(OT, \sigma_z, \sigma_r)$ (the latter accommodates the observed input ($\sigma_z$) / output ($\sigma_r$) behavior as defined by the $L^*$ algorithm).
 In Step 2, procedure buildRewardMa-\\chine$(OT)$ constructs the new hypothesized MRM $\mathcal{H}$.
 Procedures getMQ(), resolve-\\MQ() and addCounterExample() are part of the $L^*$ algorithm and are assume to be predefined.
 In our implementation, $Plan(P)$ is performed by the STORM model-checker.

\subsection{Answering Membership Queries Efficiently}
\label{sec:Answering-Membership-Queries-Efficiently}
\noindent Execution of $L^*$ requires answers to membership queries: given a sequence of observations $\sigma_z$, we need to discover what the associated reward $\sigma_r=\mathcal{H}(\sigma_z)$ is. Because the transitions in the MDP are stochastic, there is in general no sequence of actions $\bar{a}$ that is guaranteed to force the occurrence of $\sigma_z$. Forcing the sequence $\sigma_z$ is a {\em planning} problem. Procedure getExperience$(\sigma_z, s_0)$ implements this in the algorithm formalizing the framework.
Here we distinguish two natural variants of this planning problem.

The first variant, called {\it MAX} asks to synthesize the strategy $\pi_{\it MAX}$ that produces (from the initial state of the MDP) the sequence $\sigma_z$ with the {\em highest probability} $\alpha$. Let $\pi_{{\it MAX}}$ be the optimal strategy for {\it MAX}. Then we can play $\pi_{{\it MAX}}$ repeatedly, and reset the system as soon as the prefix of observations is not compatible with $\sigma_z$. Each such try will succeed with probability $\alpha$, and so by repeating trials, we are guaranteed to observe $\sigma_z$ with probability $1-(1-\alpha)^n$ after $n$ trials, and so with probability one in the long run. To synthesize the strategy $\pi_{{\it MAX}}$, we construct a MDP $M'$ from $M$ and $\sigma_z$ in which states are pairs $(s,i)$ where $s$ is a state of $M$ and $i$ is an index that tracks the prefix of the sequence $\sigma_z$ that has been observed so far. MDP $M'$ is reset as soon as the observed prefix is not compatible with $\sigma_z$. The goal is to reach any state $(s,k)$ where $k$ is the length of the sequence of observations $\sigma_z$. The model-checking tool STORM can then be used to synthesized, in polynomial time, the strategy that maximizes the probability $\alpha$ of reaching this set of states.

The second variant, called {\it MIN} is more refined and asks to synthesize the strategy $\pi_{\it MIN}$ that minimize the {\em expected number of steps} $N$ that are needed (including resets of $M$) in order to observe $\sigma_z$. This strategy $\pi_{\it MIN}$ is played and the sequence of observations $\sigma_z$ is observed after an expected number of $N$ steps. Again, we use STORM to compute this strategy in polynomial time. It can be computed as the strategy that minimizes the expected cost (each step costs $1$) to reach a state of the form $(s,k)$ where $k$ is the length of the sequence of observations $\sigma_z$ in the MDP $M'$ described in the previous paragraph.

\section{\uppercase{Formal Guarantees of the Learning\\ Framework}}
\label{sec:Guarantees}

\noindent Our framework provably offers two guarantees that can be stated as follows: ${\bf (1)}$ When playing $\pi^*_\mathcal{H}$ online, with probability 1: either we obtain $V(\pi^*_\mathcal{H})$ in the long-run or we observe a counter-example to $\mathcal{H}$, ${\bf (2)}$ If the expert's value, $V_\mathit{expert}$ is less than the long-run value of playing the optimal strategy (of the underlying reward machine $\mathcal{R}$), then with probability 1, we learn hypothesis $\mathcal{H}$ such that $V(\pi^*_\mathcal{H}) \geq V_\mathit{expert}$. In the rest of this section, we prove these two claims.

\subsection{Consequence of playing $\pi^*_\mathcal{H}$ on $M \otimes_\lambda^\curvearrowleft \mathcal{R}$}
\noindent We introduce a special \textit{counter-example} state $CE$ for the theory in this section. $CE$ is entered when a counter-example is encountered.
To describe the effect of playing $\pi^*_\mathcal{H}$ online, we need to work in the state-space $(S\times U^\mathcal{R}\times U^\mathcal{H}) \cup \{CE\}$ of process $M \otimes_\lambda^\curvearrowleft \mathcal{R},\mathcal{H}$ (defined below) and to study the associated Markov chain when $\pi^*_\mathcal{H}$ is used.
Intuitively, we use $s$ and $u_\mathcal{H}$ in $\langle s, u_\mathcal{R},u_\mathcal{H} \rangle$ to determine what action is played by $\pi^*_\mathcal{H}$. The reward observed online from $\mathcal{R}$ and predicted by $\mathcal{H}$ are compared: either the rewards for $u_\mathcal{R}$ and $u_\mathcal{H}$ in the two RMs agree and the RMs are updated accordingly, or we go to a special \textit{counter-example} state $CE$.

Formally, the transition probabilities between states in this space are defined as follows. Let $a\in A\setminus\{\curvearrowleft\}$ and $z=\lambda(a,s')$ then:



\[
T(\langle s, u_\mathcal{R},u_\mathcal{H} \rangle, a, \langle s', u_\mathcal{R}',u_\mathcal{H}' \rangle)=
\left\{
\begin{array}{ll}
    T(s,a,s') & \mbox{if } \left\{\begin{array}{c}
        \delta_r^\mathcal{R}(u_\mathcal{R},z) = \delta_r^\mathcal{H}(u_\mathcal{H},z) \land\\ \delta_u^\mathcal{R}(u_\mathcal{R},z) = u_\mathcal{R}' \land\\
        \delta_u^\mathcal{H}(u_\mathcal{H},z) = u_\mathcal{H}'
    \end{array}{}\right.\\
    0 & \mbox{otherwise.}
\end{array}{}
\right.
\]

\[
T(\langle s, u_\mathcal{R},u_\mathcal{H} \rangle, a, {\it CE}) =
\sum_{s ' \mid \delta_r^\mathcal{R}(u_\mathcal{R},z) \neq \delta_r^\mathcal{H}(u_\mathcal{H},z)} T (s,a,s')\]


\[
T(\langle s, u_\mathcal{R},u_\mathcal{H} \rangle, \curvearrowleft, \langle s', u_\mathcal{R}',u_\mathcal{H}' \rangle) =
\left\{
\begin{array}{ll}
    1 & \mbox{if } \langle s', u_\mathcal{R}',u_\mathcal{H}' \rangle = \langle s_0, u_{\mathcal{R}0},u_{\mathcal{H}0} \rangle\\
    0 & \mbox{otherwise.}
\end{array}{}
\right.    
\]

Additionally, ${\it CE}$ is a sink (never left when entered). Reaching ${\it CE}$ occurs when a counter-example to the equivalence between $\mathcal{R}$ and $\mathcal{H}$ has been discovered online. If $h$ is the history that reaches ${\it CE}$, then $h$ can be used to restart $L^*$ and compute a new hypothesis $\mathcal{H}'$ as it contains a sequence of observations on which $\mathcal{R}$ and $\mathcal{H}$ disagree.
%


\begin{definition} [BSCC \cite{bk08b}]
Let $M = \langle S, A, T, s_0 \rangle$ be an nrMDP and $C \subseteq S$.
Let $P(s,s')$ be the probability of reaching state $s'$ from state $s$ (via a sequence of actions).
$C$ is {\em strongly connected} if for every pair of states $v, w \in C$, the states $v$ and $w$ are mutually reachable, i.e., $P(v,w)>0$ and $P(w,v)>0$.
A {\em strongly connected component} (SCC) of $M$ is a maximally strongly connected set of states. That is, $C$ is an SCC if $C$ is strongly connected and $C$ is not contained in another strongly connected set of states $D\subseteq S$ with $C \neq D$. A {\em bottom} SCC (BSCC) of $M$ is an SCC $B$ from which no state outside $B$ is reachable, i.e., for each state $b \in B$ it holds that $\sum_{b'\in B}P(b,b') = 1$.
\end{definition}

\begin{theorem}
\label{th:1}
When playing $\pi^*_\mathcal{H}$ online, with probability 1: either $V(\pi^*_\mathcal{H})$ is obtained or a counter-example to $\mathcal{H}$ is found.
\end{theorem}
\noindent {\bf Proof}
To prove this theorem, we show that if no counter-example is found when playing $\pi^*_\mathcal{H}$ online (playing on $M \otimes_\lambda^\curvearrowleft \mathcal{R},\mathcal{H}$), then the long run mean-payoff of the observed outcome is $V(\pi^*_\mathcal{H})$ with probability one.

First, we note that the MDP $M \otimes_\lambda^\curvearrowleft \mathcal{H}$ on which we compute $\pi^*_\mathcal{H}$ is strongly connected. As a consequence, all the states of the MDP have the same optimal mean-payoff value which is $V(\pi^*_\mathcal{H})$. It also means that all the states in the associated Markov chain, noted $M \otimes_\lambda^\curvearrowleft \mathcal{H}(\pi^*_\mathcal{H})$, have the same value $V(\pi^*_\mathcal{H})$. In turn this implies that all the BSCC of this Markov chain have the same value $V(\pi^*_\mathcal{H})$. From that, we deduce that the mean-payoff obtained by a random walk in those BSCCs has a value equal to $V(\pi^*_\mathcal{H})$ with probability one (see e.g.~\cite{p94b}).

Next, we are interested in the expected value obtained when playing $\pi^*_\mathcal{H}$ on $M \otimes_\lambda^\curvearrowleft \mathcal{R},\mathcal{H}$ conditional to the fact that no counter-example is encountered, that is, while $CE$ is not reached.
We claim that the behavior of $\pi^*_\mathcal{H}$ on $M \otimes_\lambda^\curvearrowleft \mathcal{R},\mathcal{H}$ is then exactly the same as in $M \otimes_\lambda^\curvearrowleft \mathcal{H}$. This is a consequence of the following property: there are only two kinds of BSCCs in $M \otimes_\lambda^\curvearrowleft \mathcal{R},\mathcal{H} \, (\pi^*_\mathcal{H})$: (i) $\{CE\}$ and (ii) BSCCs $B$ not containing $CE$. When in $B$, $\mathcal{R}$ and $\mathcal{H}$ fully agree in the sense that, for all $\langle s, u_\mathcal{R},u_\mathcal{H} \rangle \in B$, as $CE$ cannot be reached (otherwise $B$ would not be a BSCC), the action chosen by $\pi^*_\mathcal{H}$ triggers the same reward from $\mathcal{R}$ and $\mathcal{H}$. So we conclude that the expected reward in $B$ is as predicted by $\mathcal{H}$ and is equal to $V(\pi^*_\mathcal{H})$. Hence, again the value observed will be $V(\pi^*_\mathcal{H})$ with probability $1$ and the theorem follows. \hfill$\blacksquare$

\subsection{Leveraging off Expert Knowledge}
\noindent Now we show how one can take advantage of the knowledge a domain expert has about what value can be expected when playing optimally in the domain.
This idea is exploited in Step 3 of the description of the flow of our algorithm (Sect.~\ref{sec:Flow-of-Operation}).

To this end, we first establish how we can find counter-examples systematically using random exploration. For that, we define the strategy $\pi_u$ which plays uniformly at random all actions in $A\cup \{\curvearrowleft\}$, that is,
\[
\pi_u : S\times U^\mathcal{R}\times U^\mathcal{H} \mapsto \Delta(A\cup \{\curvearrowleft\})
\]
such that $\pi_u(\langle s, u_\mathcal{R},u_\mathcal{H} \rangle)(a) = 1/(|A|+1)$.\footnote{$\Delta(X)$ is all probability distributions over set $X$.}
\noindent
and prove the following result.
\begin{lemma}
\label{lm:1}
While playing $\pi_u$ online, a counter-example is found (with probability one) if and only if $\mathcal{R}$ and $\mathcal{H}$ are not equivalent.
\end{lemma}
\noindent {\bf Proof}
First, we note that the Markov chain $M \otimes_\lambda^\curvearrowleft \mathcal{R},\mathcal{H} \, (\pi_u)$ is composed of two strongly connected components: $C_1=\{CE\}$ and $C_2$ that contains all the other states. We also note that 
$C_1=\{CE\}$ is reachable in this Markov chain iff $\mathcal{R}$ and $\mathcal{H}$ are not equivalent.
So, when playing $\pi_u$ on $M \otimes_\lambda^\curvearrowleft \mathcal{R},\mathcal{H}$, two mutually exclusive scenarios are possible. First, $\mathcal{R}$ and $\mathcal{H}$ are not equivalent. Then, there are transitions from $C_2$ to $C_1$, and $C_1$ is the only BSCC of the Markov chain. As a consequence, the execution of $\pi_u$ ends in $C_1$ with probability 1. Second, $\mathcal{R}$ and $\mathcal{H}$ are equivalent. Then, there are no transitions from $C_2$ to $C_1$, and $C_2$ is the only reachable BSCC of the Markov chain and no counter-examples exist. \hfill$\blacksquare$

Suppose that a domain expert can provide us with a value $V_{\it expert}$ which is realistic in the following sense: $V_{\it expert}$ is below the {\em true} optimal value achievable on $M \otimes_\lambda^\curvearrowleft \mathcal{R}$. Then the following theorem holds.
\begin{theorem}
If $V_\mathit{expert}$ is less than or equal to the optimal value $V(\pi^*_\mathcal{R})$ of $M \otimes_\lambda^\curvearrowleft \mathcal{R}$, then with probability 1, a hypothesis $\mathcal{H}$ will be learned such that $V(\pi^*_\mathcal{H}) \geq V_\mathit{expert}$.
\end{theorem}
\noindent {\bf Proof}
Note that if $V(\pi^*_\mathcal{H}) < V_\mathit{expert}$, then it must be the case that the hypothesized MRM $\mathcal{H}$ does not model the underlying MRM $\mathcal{R}$ (i.e., they are not equivalent). Furthermore, if $V(\pi^*_\mathcal{H}) < V_\mathit{expert}$, then our algorithm plays the uniform random strategy $\pi_u$, and by Lemma~\ref{lm:1}, a counter-example is found with probability 1 which triggers $L^*$ to be restarted in order to obtain a new hypothesis $\mathcal{H}'$. 
Now, we note that the scenario above can happen only a number of times bounded by the number of states in $\mathcal{R}$, because the $L^*$ algorithm is guaranteed to learn $\mathcal{R}$ after a number of equivalence queries that is at most equal to the number of states in $\mathcal{R}$~\cite{a87}. So, either we find a hypothesis that predicts a better value than $V_\mathit{expert}$, or we end up with an hypothesis $\mathcal{H}$ which is correct and so equivalent to $\mathcal{R}$, implying that $V(\pi^*_\mathcal{H}) = V(\pi^*_\mathcal{R}) \geq V_\mathit{expert}$. We conclude that our algorithm is guaranteed to obtain online a mean-payoff value which is at least $V_\mathit{expert}$ with probability $1$. \hfill$\blacksquare$

\section{\uppercase{Experimental Evaluation}}
\label{sec:Experimental-Evaluation}

\noindent Our online active-learning algorithm
was implemented, and evaluated on three small problem domains.\footnote{For the $L^*$ algorithm, we used a Python implementation provided by Georges Bossert (https://github.com/gbossert/pylstar).}
The three problems are also evaluated w.r.t.\ a baseline based on neural networks. In this section, we describe the baseline, and then in the following three subsections, we describe the problem domain, describe in more detail the experiment setup for the particular problem and present the results. However, some aspects of the experiments are common to all our cases and discussed immediately:

For convenience, we refer to our framework as ARM (for Angluin Reward Machine) and the baseline as DQN (for Deep Q-function Network).
When using ARM, the number of actions allowed per trial includes actions required for answering membership queries and for searching for counter-examples. Rewards received during exploration, learning, exploitation are all considered.
In all domains, the agent can move north, east, west and south, one cell per action. To add stochasticity to the transition function, the agent is made to get stuck $5\%$ of the time when executing an action.
We measure the cumulative rewards gained (Return) per episode.
%
Procedure getExperience is set to use mode {\it MIN} (cf.\ Sect.~\ref{sec:Answering-Membership-Queries-Efficiently}) in all experiments.
Episode length, total number of steps (actions) and number of trials performed are domain dependent; they will be mentioned for each domain.
To achieve similar total experiment times (per problem), episode lengths of DQN were increased.

\subsection{Deep Q-learning Baseline}

We implemented a Deep Q-learning network agent (DQN) \cite{mksrvbgrfo15}. To improve the stability and the convergence speed of the learning, we augmented the agent with two standard techniques: experience replay \cite{mksrvbgrfo15} and double-q function \cite{h10}.

Let $f_Q: \hat{S} \to \mathbb{R}^{|A|}$ be a standard deep Q function, receiving as input a state representation and returning as output the real vector of Q-values for each single action. The state representation space $\hat{S}$ depends on the particular domain. For the domains used in this paper, it is the set of $(x,y)$ coordinates of the grid. 
 Since `regular' deep Q-learning algorithms assume a Markovian setting, a naive application of deep reinforcement learning techniques does not provide us with a fair competitor. Therefore, we enhanced our DQN baseline with the ability to make decisions based on observation history:
%
 In particular, we extended the standard deep Q network in three ways. 
 
 First, we added a one-hot representation of the \textit{current} observation to the input space. Let $i$ be the index of the current observation, its one-hot representation is a zero-vector with a single $1$ in position $i$. For example if the observation $map$ is indexed as $i=2$ out of $4$ possible observations, its one-hot representation is $v = [0,1,0,0]$. Let $\hat{O}$ be the space of one-hot representations of the current observation.
 
 Second, we added a bag-of-words (BoW) representation of the \textit{past} observations. Given a sequence of past observations, its BoW representation is a vector $v$ representing the multiset of the observations, i.e. $v_i$ is equal to the number of times observation indexed by $i$ is encountered in the history. For example, if the past observations were $\langle map, treasure, map\rangle$ (in that order), indexed by $I =[2,4,2]$, then their BoW representation is $h = [0,2,0,1]$.
 Note that if the agent sees $\langle treasure, map, map \rangle$ (in that order), then the representation of the history is still $h = [0,2,0,1]$. Hence, a Q-learning agent can distinguish bags of observations, but not the order in which they were perceived. Sometimes BoW is a sufficient statistics for discriminating between cases, like in document classification \cite{mcg16}.\footnote{Our intention is to provide a reasonable baseline, not to develop an excellent neural-network-based solution for non-Markovian domains.}
 Let $\hat{H}$ be the set of BoW representations of the past observations. 
 
 Third, we extended the action set $A$ with a \textit{reset} action that mimics the behavior of $\curvearrowleft$ in our ARM framework. This is resembling the way our algorithm interacts with the environment. The reset action is a special one, because the agent is forced to reset its past observation representation (i.e. set to the zero-vector) each time this action is selected.
 
 Finally, we can define the history-augmented deep Q function as the function $f^h_Q: \hat{S} \times \hat{O} \times \hat{H} \to \mathbb{R}^{|A|+1}$.
 
 We implemented the deep network and the Q-learning algorithm in Tensorflow \cite{abccdddgii16}. Once the input space of the network is augmented with the history-based information, standard RL frameworks can be exploited. In particular, we selected KerasRL \cite{p16}. The neural network architecture is a feed-forward neural network, with three hidden layers with fifty ReLU units each and a linear output layer. The Adam algorithm \cite{kb14} with a learning rate of $10^{-3}$ is selected as weight-update rule.The epsilon-greedy policy is used, with $\epsilon=0.1$.

\subsection{Treasure-Map World}
\begin{figure}
\centering
\includegraphics[scale=0.4]{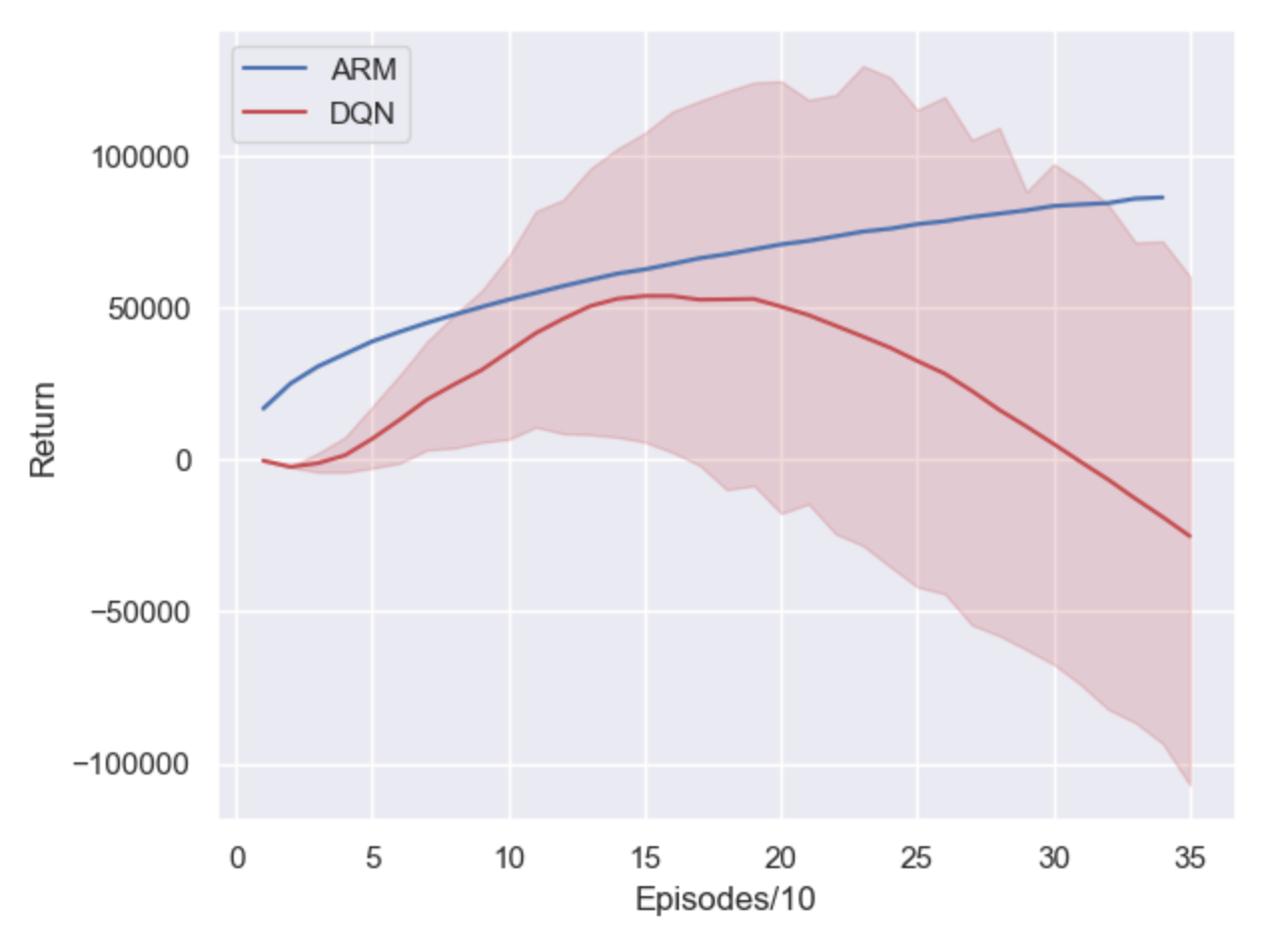}
\caption{Results for the Treasure-Map world (averaged over ten trials for ARM and twenty trials for DQN). Episodes were 507 steps long for ARM and 1014 steps for DQN. Average running-times per episode: ARM 24.9s; DQN: 10.4s. 
\label{fig:tmw-results}}
\end{figure}

\noindent The agent starts in one of the four corners of one of the five areas; one of these twenty locations is randomly chosen.
The default cost $c$ was set to -0.1 and the cost for resetting was set to -10.
The optimal value for this domain is 9.884 (computed by STORM for the correct reward machine). We set $V_\mathit{expert} = 9$.
The performance of our framework applied to the Treasure-Map world can be seen in Figure~\ref{fig:tmw-results}.
On average, for ARM, there were 835 membership-queries, and 509 counter-examples found during exploitation.

\subsection{Office-Bot Domain}

\begin{figure}
\centering
\begin{subfigure}[b]{.35\textwidth}
\includegraphics[width=1\linewidth]{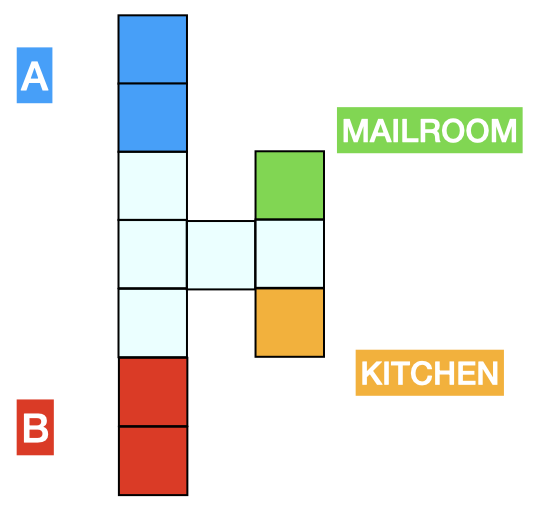}  
  \caption{The layout of the office.}
  \label{fig:obd-layout}
\end{subfigure}
\hspace{4mm}
\begin{subfigure}[b]{.5\textwidth}
\centering
\vspace{5mm}
\begin{tikzpicture}[->,>=stealth',shorten >=1pt,auto,node distance=3cm,
  thick,main node/.style={circle,fill=blue!10,draw,
  font=\sffamily\normalsize\bfseries, minimum size=8mm, scale=0.6, transform shape}]

  \node[main node] (0) {0};
  \node[main node] (1) at (2,0.75) {1};
  \node[main node] (2) at (2,0.25) {2};
  \node[main node] (3) at (2,-0.25) {3};
  \node[main node] (4) at (2,-0.75) {4};
  \node[main node] (5) at (4,0.75) {5};
  \node[main node] (6) at (4,0.25) {6};
  \node[main node] (7) at (4,-0.25) {7};
  \node[main node] (8) at (4,-0.75) {8};
  \coordinate (x) at (6,0);
  \coordinate (y) at (4,-2);
  
  \path[every node/.style={sloped,anchor=south,auto=false, font=\sffamily\tiny,fill=none, inner sep=1pt}]
  	(0) edge [] node[near end] {$mrA\mid1$} (1)
    (0) edge [] node[near end] {$mrB\mid1$} (2)
    (0) edge [] node[near end] {$drA\mid1$} (3)
    (0) edge [] node[near end] {$drB\mid1$} (4)

    (1) edge [] node[] {$hmA\mid2$} (5)
    (2) edge [] node[] {$hmB\mid2$} (6)
    (3) edge [] node[] {$hdA\mid2$} (7)
    (4) edge [] node[] {$hdB\mid2$} (8)

    (5) edge [style={-}] node[near start] {$del\mid3$} (x)
    (6) edge [style={-}] node[near start] {$del\mid3$} (x)
    (7) edge [style={-}] node[near start] {$del\mid4$} (x)
    (8) edge [style={-}] node[near start] {$del\mid4$} (x)

    (x) edge [style={-}, inner sep=0pt, out=0, in=0] (y)
    (y) edge [out=180,in=-90] node[] {} (0);
  \end{tikzpicture}
  \vspace{0mm}
\caption{A Mealy reward machine. Self-loops are not shown.}
  \label{fig:obd-MRM}
\end{subfigure}
\vspace{2mm}
\caption{The Office-bot Domain.}
\label{fig:office-bot-scenario}
\end{figure}
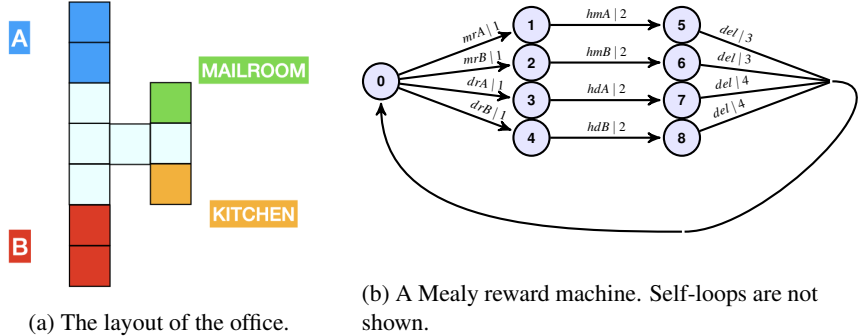

\noindent We now consider a slightly more complicated problem than the Treasure-map world.
In this domain, a robot must deliver the correct mail to a person in office A or office B, or the robot must deliver a doughnut to one of those two offices. But these deliveries must be initiated by a request from an office occupant. Mail is collected in the mail room and doughnuts in the kitchen. What makes this a non-Markovian situation is that the robot must remember who asked for an item and what item was asked for. The office layout and the underlying reward machine are shown in Figure~\ref{fig:office-bot-scenario}.

Besides the four movement actions, there are actions $\mathtt{ask}$ (ask what occupant wants), $\mathtt{pickMailA}$ (pickup mail for A), $\mathtt{pickDonutA}$, $\mathtt{dropItemA}$ (drop item at A), $\mathtt{pickMailB}$, $\mathtt{pickDonutB}$, and $\mathtt{dropItemB}$.

The observation that the robot can make are
$\mathtt{mrA}$ (mail request by A), $\mathtt{drA}$ (doughnut request by A), $\mathtt{hmA}$ (have mail for A), $\mathtt{hdA}$ (have doughnut for A), $\mathtt{mrB}$ (mail request by B), $\mathtt{drB}$ (doughnut request by B), $\mathtt{hmB}$ (have mail for B), $\mathtt{hdB}$ (have doughnut for B), $\mathtt{del}$ (item delivered).

The robot gets 1 reward point for making a request, 2 points for having one of the two items in its possession, 3 points for delivering a doughnut to the correct person, and 4 points for delivering mail to the correct person. We define ``correct person'' via the labeling function by defining, for instance, $\lambda(\mathtt{dropItemA},s_A)=\mathtt{del}$, where $s_A$ is a state representing A's office (one of the two blue cells).

When the bot asks someone in an office what they want, the bot gets an answer \textit{doughnut} or \textit{mail} with $50\%$ chance each. Our framework cannot deal with stochastic observations, so we encoded this behavior into the transition function: Whenever the bot is in an office and asks, it transitions to one of the two cells of the office with $50\%$ probability. Then, if the bot is (ends up) in the back of the office, it is told to fetch mail, else it is told to fetch a doughnut. For instance, for action $\mathtt{ask}$, the labeling function simply produces $\mathtt{mrB}$ (mail request by B) in the bottom red cell, and $\mathtt{drB}$ (doughnut request by B) in the top red cell.

\begin{figure}
\centering
\includegraphics[scale=0.4]{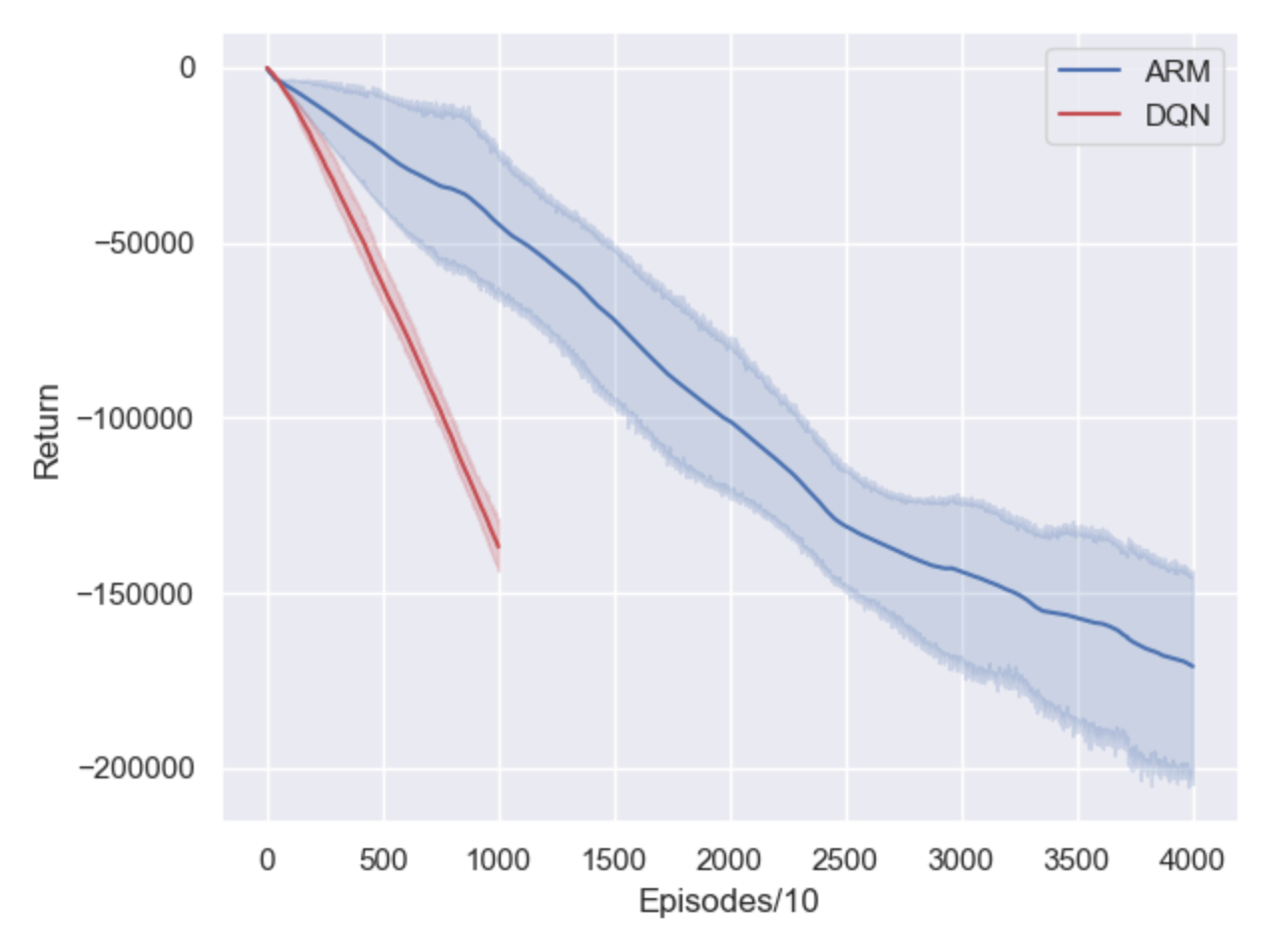}
\caption{Results for the Office-Bot domain (averaged over ten trials for ARM and five trials for DQN). Episodes were 63 steps long. Significantly fewer episodes were run for DQN, because running-time is much longer (Avg. running-times per episode: ARM: 0.091s; DQN: 0.462s).
\label{fig:obd-results}}
\end{figure}

The robot starts in the middle of the hallway.
The default cost $c$ was set to -0.1 and the cost for resetting was set to -2.
The optimal value for this domain is 0.383 (computed by STORM for the correct reward machine). We set $V_\mathit{expert} = 0.3$.
The performance of our framework applied to the Office-Bot domain can be seen in Figure~\ref{fig:obd-results}. 
On average, for ARM, there were 6060 membership-queries, and 152 counter-examples found during exploitation.
Performing more than 10000 episodes with DQN would have taken too long.

The MRM learnt by ARM is actually more compact than the one provided to be learned; the learnt one is like the MRM in Figure \ref{fig:obd-MRM}, but with nodes 5 and 6, resp., 7 and 8 merged. Angluin's algorithm always finds the minimal machine.

\subsection{The Cube Problem}
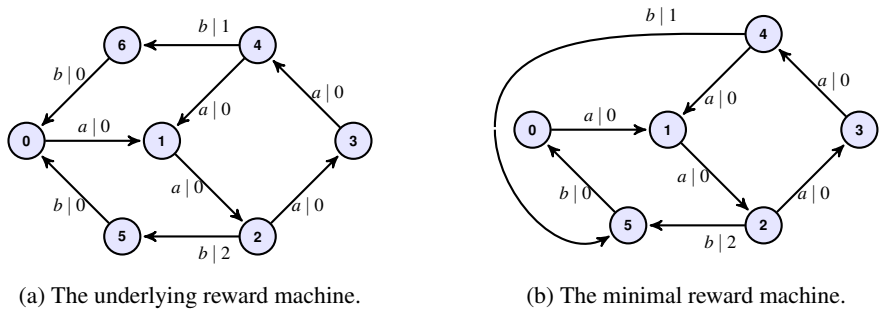
\begin{figure}
\begin{subfigure}[b]{.5\textwidth}
\centering
\begin{tikzpicture}[->,>=stealth',shorten >=1pt,auto,node distance=3cm,
  thick,main node/.style={circle,fill=blue!10,draw,
  font=\sffamily\normalsize\bfseries,minimum size=8mm, scale=0.6, transform shape}]

  \node[main node] (0) {0};
  \node[main node] (1) [right of=0] {1};
  \node[main node] (2) [below right of=1] {2};
  \node[main node] (3) [above right of=2] {3};
  \node[main node] (4) [above left of=3] {4};
  \node[main node] (5) [below right of=0] {5};
  \node[main node] (6) [above right of=0] {6};
  
  \path[every node/.style={near start, font=\sffamily\scriptsize,fill=none,inner sep=1pt}]
    (0) edge [] node[above, midway] {$a\mid0$} (1)
    (1) edge [] node[left=1pt, midway] {$a\mid0$} (2)
    (2) edge [] node[right] {$a\mid0$} (3)
    edge [] node[below=2pt] {$b\mid2$} (5)
    (3) edge [] node[right=1pt, midway] {$a\mid0$} (4)
    (4) edge [] node[right=1pt, near end] {$a\mid0$} (1)
    edge [] node[above=2pt, near start] {$b\mid1$} (6)
    (5) edge [] node[left=1pt] {$b\mid0$} (0)
    (6) edge [] node[left=1pt] {$b\mid0$} (0);
\end{tikzpicture}
\caption{The underlying reward machine.}
\end{subfigure}
\hspace{4mm}
\begin{subfigure}[b]{.5\textwidth}
\centering
\vspace{2mm}
\begin{tikzpicture}[->,>=stealth', shorten >=1pt, auto, node distance=3cm, 
  thick, main node/.style={circle,fill=blue!10, draw,
  font=\sffamily\normalsize\bfseries,minimum size=8mm, scale=0.6, transform shape}]

  \node[main node] (0) {0};
  \node[main node] (1) [right of=0] {1};
  \node[main node] (2) [below right of=1] {2};
  \node[main node] (3) [above right of=2] {3};
  \node[main node] (4) [above left of=3] {4};
  \node[main node] (5) [below right of=0] {5};
  \coordinate (x) at (-0.5,0);
  
  \path[every node/.style={near start, font=\sffamily\scriptsize,fill=none,inner sep=1pt}]
    (0) edge [] node[above, midway] {$a\mid0$} (1)
    (1) edge [] node[left=1pt, midway] {$a\mid0$} (2)
    (2) edge [] node[right] {$a\mid0$} (3)
    edge [] node[below=2pt] {$b\mid2$} (5)
    (3) edge [] node[right=1pt, midway] {$a\mid0$} (4)
    (4) edge [] node[right=1pt, near end] {$a\mid0$} (1)
    edge [style={-},inner sep=0pt,out=180,in=90] node[above=2pt, near start] {$b\mid1$} (x)
    (x) edge [out=-90,in=210] (5)
    (5) edge [] node[left=1pt] {$b\mid0$} (0);
\end{tikzpicture}
\caption{The minimal reward machine.}
\end{subfigure}
\vspace{2mm}
\caption{The cube reward machine.
\label{fig:cube-underlying}}
\end{figure}

\noindent This problem is added for more variety.
The environment is a 5-by-5 grid without obstacles, and with two cells where $a$ is perceived and two cells where $b$ is perceived. The underlying reward model is the one shown in Figure~\ref{fig:cube-underlying} (top).

The agent always starts in the bottom right-hand corner.
The default cost $c$ was set to 0 and the cost for resetting was set to -1.

The optimal value for this problem is 0.1624 (computed by STORM for the correct reward machine). We set $V_\mathit{expert} = 0.15$.
If $V_\mathit{expert}$ is set too high, then the agent will keep on seeking counter-examples.
Initially (for the first hypothesis RM), STORM computes an optimal value below 0.15. Random exploration commences, a counter-example is soon found and a correct reward machine is learnt.

\begin{figure}
\centering
\includegraphics[scale=0.4]{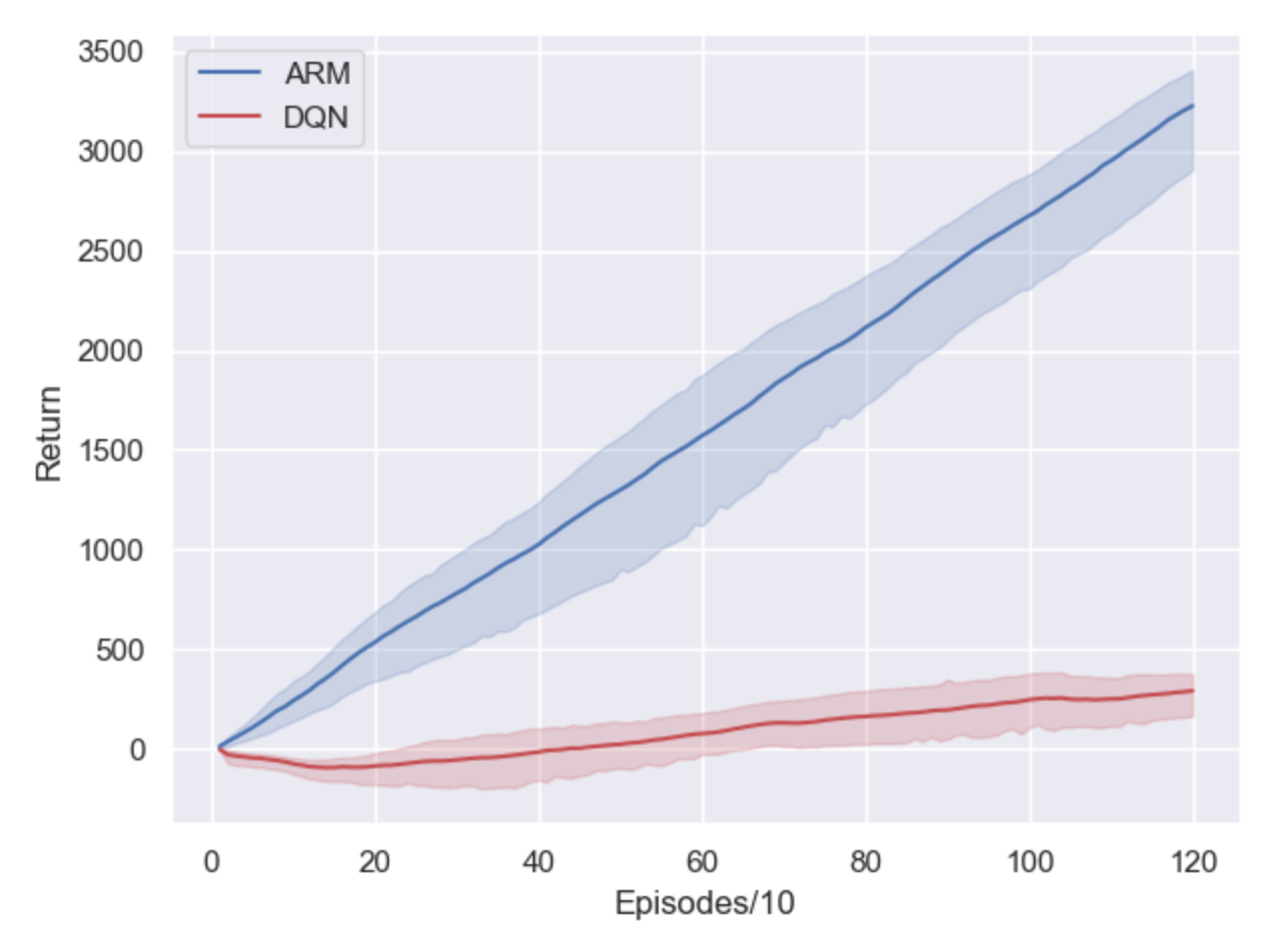}
\caption{Results for the Cube problem (averaged over five trials). Episodes were 75 steps long for ARM and 600 steps for DQN. Average running-times per episode: ARM: 1.1s; DQN: 45s.
\label{fig:cube-results}}
\end{figure}

The performance of our framework applied to the Cube problem can be seen in Figure~\ref{fig:cube-results}.
On average, for ARM, there were 850 membership-queries, and 248 counter-examples found during exploitation.

In this example (and in the Office-Bot domain), another benefit of the $L^*$ algorithm can be seen: it produces the minimal machine (Figure~\ref{fig:cube-underlying}, bottom), which usually makes its meaning clearer, and produces exponentially smaller synchronized product MDPs.

\section{\uppercase{Conclusion}}

\noindent We proposed a framework for learning and exploiting non-Markovian reward models, represented as Mealy machines. Angluin's $L^*$ algorithm is employed to learn \textit{Mealy Reward Machines} within a Markov Decision Process setting. The framework was implemented and used for evaluation as a proof-of-concept.

Some useful theorems were proven, relying on existing theory of probabilistic model-checking and automata learning theory. The main result is that, under reasonable assumptions, we can guarantee that a system will eventually achieve a performance level provided by a domain expert if that performance level is realistic/achievable for that domain.

We found that in empirical evaluation, the framework learns the underlying MRM of the applicable domain correctly, that is, after answering a finite number of membership queries as posed by the $L^*$ algorithm, within a finite time. And it significantly outperforms a Deep Q-Network augmented to take history-based observations as input. Moreover, the learnt MRM is minimal; least possible number of nodes.

As investigated in some of the related literature mentioned in the introduction, an agent with a non-Markovian (reward) model is better equipped to avoid dangerous or undesirable situation in some domains. A challenge is, how to safely \textit{learn} non-Markovian (reward) models in unsafe domains? Some work has been done on safe learning/exploration \cite{tbk16,hka19,comb19}

We expect that more complex non-Markovian reward environments will require intelligent exploration strategies.
It will be interesting to investigate the exploration-exploitation trade-off in the non-Markovian setting.
And if an expert value is not available, one could employ an $\epsilon$-greedy (or similar) strategy to allow for the agent to observe counter-examples if they exist. This will be investigated in future work.

\bibliography{references}

\begin{thebibliography}{10}

\bibitem{abccdddgii16}
M.~Abadi, P.~Barham, J.~Chen, Z.~Chen, A.~Davis, J.~Dean, M.~Devin,
  S.~Ghemawat, G.~Irving, M.~Isard, et~al., `Tensorflow: A system for
  large-scale machine learning', in {\em Twelfth {USENIX} symposium on
  operating systems design and implementation ({OSDI} 16)}, pp. 265--283,
  (2016).

\bibitem{abeknt18}
M.~Alshiekh, R.~Bloem, R.~Ehlers, B.~K\"onighofer, S.~Niekum, and U.~Topcu,
  `Safe reinforcement learning via shielding', in {\em Proceedings of the
  Thirty-Second {AAAI} Conf. on Artif. Intell. (AAAI-18)}, pp. 2669--2678. AAAI
  Press, (2018).

\bibitem{abz10}
C.~Amato, B.~Bonet, and S.~Zilberstein, `Finite-state controllers based on
  {Mealy} machines for centralized and decentralized {POMDPs}', in {\em
  Proceedings of the Twenty-Fourth AAAI Conf. on Artif. Intell. (AAAI-10)}, pp.
  1052--1058. AAAI Press, (2010).

\bibitem{a87}
D.~Angluin, `Learning regular sets from queries and counterexamples', {\em
  Information and Computation}, {\bf 75}(2),  87--106, (1987).

\bibitem{bbg96}
F.~Bacchus, C.~Boutilier, and A.~Grove, `Rewarding behaviors', in {\em
  Proceedings of the Thirteenth Natl. Conf. on Artif. Intell.}, pp. 1160--1167.
  AAAI Press, (1996).

\bibitem{bk08b}
C.~Baier and J.{-}P. Katoen, {\em Principles of Model Checking}, {MIT} Press,
  2008.

\bibitem{bdp18}
R.~Brafman, G.~De Giacomo, and F.~Patrizi, `{LTL$_f$/LDL$_f$} non-{M}arkovian
  rewards', in {\em Proceedings of the The Thirty-Second AAAI Conference on
  Artificial Intelligence (AAAI-18)}, pp. 1771--1778. AAAI Press, (2018).

\bibitem{ccsm18}
A.~Camacho, O.~Chen, S.~Sanner, and S.~McIlraith, `Non-{M}arkovian rewards
  expressed in {LTL}: Guiding search via reward shaping (extended version)', in
  {\em Proceedings of the First Workshop on Goal Specifications for
  Reinforcement Learning}, FAIM 2018, (2018).

\bibitem{ctkvm19}
A.~Camacho, R.~Toro Icarte, T.~Klassen, R.~Valenzano, and S.~McIlraith, `{LTL}
  and beyond: Formal languages for reward function specification in
  reinforcement learning', in {\em Proceedings of the Twenty-Eighth
  International Joint Conference on Artificial Intelligence}, IJCAI-19, pp.
  6065--6073, (2019).

\bibitem{comb19}
R.~Cheng, G.~Orosz, R.~Murray, and J.~Burdick, `End-to-end safe reinforcement
  learning through barrier functions for safety-critical continuous control
  tasks', in {\em The Thirty-third {AAAI} Conference on Artificial
  Intelligence}, pp. 3387--3395. {AAAI} Press, (2019).

\bibitem{dfip19}
G.~De Giacomo, M.~Favorito, L.~Iocchi, and F.~Patrizi, `Foundations for
  restraining bolts: Reinforcement learning with {LTL$_f$/LDL$_f$} restraining
  specifications', in {\em Proceedings of the Twenty-Ninth International
  Conference on Automated Planning and Scheduling (ICAPS-19)}, pp. 128--136.
  AAAI Press, (2019).

\bibitem{hak19}
M.~Hasanbeig, A.~Abate, and D.~Kroening, `Logically-constrained neural fitted
  q-iteration', in {\em Proceedings of the Eighteenth Intl. Conf. on Autonomous
  Agents and Multiagent Systems}, eds., N.~Agmon, M.~E. Taylor, E.~Elkind, and
  M.~Veloso, AAMAS-2019, pp. 2012--2014. International Foundation for AAMAS,
  (2019).

\bibitem{hka19}
M.~Hasanbeig, D.~Kroening, and A.~Abate, `Towards verifiable and safe
  model-free reinforcement learning', in {\em Proceedings of the First Workshop
  on Artificial Intelligence and Formal Verification, Logics, Automata and
  Synthesis (OVERLAY)}, (2019).

\bibitem{h10}
H.~Hasselt, `Double q-learning', in {\em Advances in neural information
  processing systems 23}, pp. 2613--2621, (2010).

\bibitem{tkvm18a}
R.~Toro Icarte, T.~Klassen, R.~Valenzano, and S.~McIlraith, `Teaching multiple
  tasks to an {RL} agent using {LTL}', in {\em Proceedings of the Seventeenth
  Intl. Conf. on Autonomous Agents and Multiagent Systems}, AAMAS-2018, pp.
  452--461. International Foundation for AAMAS, (2018).

\bibitem{tkvm18b}
R.~Toro Icarte, T.~Klassen, R.~Valenzano, and S.~McIlraith, `Using reward
  machines for high-level task specification and decomposition in reinforcement
  learning', in {\em Proceedings of the Thirty-Fifth Intl. Conf. on Machine
  Learning}, volume~80 of {\em ICML-18}, pp. 2107--2116. Proceedings of Machine
  Learning Research, (2018).

\bibitem{twkvcm19b}
R.~Toro Icarte, E.~Waldie, T.~Klassen, R.~Valenzano, M.~Castro, and
  S.~McIlraith, `Learning reward machines for partially observable
  reinforcement learning', in {\em Proceedings of the Thirty-third Conference
  on Neural Information Processing Systems}, NeurIPS 2019, (2019).

\bibitem{kb14}
D.~Kingma and J.~Ba, `Adam: A method for stochastic optimization', {\em arXiv
  preprint arXiv:1412.6980}, (2014).

\bibitem{kpr18}
J.~K\v{r}et\'insk\'y, G.~P\'erez, and J.-F. Raskin, `Learning-based mean-payoff
  optimization in an unknown {MDP} under omega-regular constraints', in {\em
  Proceedings of the Twenty-Ninth Intl. Conf. on Concurrency Theory
  (CONCUR-18)}, pp. 1--8, Schloss Dagstuhl, Germany, (2018). Dagstuhl.

\bibitem{ly96}
D.~{Lee} and M.~{Yannakakis}, `Principles and methods of testing finite state
  machines - a survey', {\em Proceedings of the IEEE}, {\bf 84}(8),
  1090--1123, (Aug 1996).

\bibitem{mcg16}
M.~McTear, Z.~Callejas, and D.~Griol, {\em The conversational interface},
  Springer Verlag, Heidelberg, New York, Dortrecht, London, 2016.

\bibitem{mksrvbgrfo15}
V.~Mnih, K.~Kavukcuoglu, D.~Silver, A.~Rusu, J.~Veness, M.~Bellemare,
  A.~Graves, M.~Riedmiller, A.~Fidjeland, G.~Ostrovski, et~al., `Human-level
  control through deep reinforcement learning', {\em nature}, {\bf 518}(7540),
  529--533, (2015).

\bibitem{p16}
M.~Plappert.
\newblock keras-rl.
\newblock \url{https://github.com/keras-rl/keras-rl}, 2016.

\bibitem{p94b}
M.~Puterman, {\em Markov Decision Processes: {D}iscrete Dynamic Programming},
  Wiley, New York, NY, 1994.

\bibitem{sg09}
M.~Shahbaz and R.~Groz, `Inferring {M}ealy machines', in {\em Proceedings of
  the International Symposium on Formal Methods (FM-09)}, eds., A.~Cavalcanti
  and D.~Dams, number 5850 in LNCS,  207--222, Springer-Verlag, Berlin
  Heidelberg, (2009).

\bibitem{tgspk06}
S.~Thi\'ebaux, C.~Gretton, J.~Slaney, D.~Price, and F.~Kabanza,
  `Decision-theoretic planning with non-{M}arkovian rewards', {\em Artif.
  Intell. Research}, {\bf 25},  17--74, (2006).

\bibitem{tbk16}
M.~Turchetta, F.~Berkenkamp, and A.~Krause, `Safe exploration in finite markov
  decision processes with gaussian processes', in {\em Proceedings of the
  Thirtieth Conference on Neural Information Processing Systems}, NeurIPS 2016,
  (2016).

\bibitem{v17}
F.~Vaandrager, `{Model Learning}', {\em Communications of the {ACM}}, {\bf
  60}(2),  86--96, (2017).

\end{thebibliography}
\bibliographystyle{ecai}
\end{document}